\documentclass[lettersize,journal]{IEEEtran}
\usepackage{amssymb}
\usepackage{amsmath}
\usepackage[english]{babel}
\usepackage{float}
\usepackage{graphicx}
\usepackage[hidelinks]{hyperref}
\usepackage{mathtools}

\usepackage{filecontents}
\usepackage[font=footnotesize,labelfont=footnotesize]{subcaption}
\usepackage[font=footnotesize,labelfont=footnotesize]{caption}
\usepackage{comment}
\usepackage{authblk}
\usepackage{multirow}
\usepackage{xcolor}
\usepackage{algorithm}
\usepackage{algorithmic}
\usepackage{xspace}
\usepackage{hhline}

\definecolor{darkblue}{rgb}{0.15,0.15,0.55}
\definecolor{lightgrey}{rgb}{0.75,0.75,0.75}

\makeindex             

\begin{document}

\title{\LARGE \bf Reinforcement Learning for Robotic Insertion of\\Flexible Cables in Industrial Settings}
\author{Jeongwoo Park$^1$, Seabin Lee$^1$, Changmin Park$^1$, Wonjong Lee$^2$, and Changjoo Nam$^{1,*}$
\thanks{This work was supported in part by Collaborative Research Program of Nachi-Fujikoshi \& Sogang University, and the National Research Foundation of Korea (NRF) grants funded by the Korea government (MSIT) (No. 2022R1C1C1008476 and RS-2024-00338772)
$^1$Dept. of Electronic Engineering, Sogang University, Seoul, Korea. $^2$Dept. of Artificial Intelligence, Sogang University, Seoul, Korea. $^*$Corresponding author: {\tt\small cjnam@sogang.ac.kr}}
}

\maketitle
\thispagestyle{empty}

\begin{abstract}
The industrial insertion of flexible flat cables (FFCs) into receptacles presents a significant challenge owing to the need for submillimeter precision when handling the deformable cables. In manufacturing processes, FFC insertion with robotic manipulators often requires laborious human-guided trajectory generation. While Reinforcement Learning (RL) offers a solution to automate this task without modeling complex properties of FFCs, the nondeterminism caused by the deformability of FFCs requires significant efforts and time on training. Moreover, training directly in a real environment is dangerous as industrial robots move fast and possess no safety measure. We propose an RL algorithm for FFC insertion that leverages a foundation model-based real-to-sim approach to reduce the training time and eliminate the risk of physical damages to robots and surroundings. Training is done entirely in simulation, allowing for random exploration without the risk of physical damages. Sim-to-real transfer is achieved through  semantic segmentation masks which leave only those visual features relevant to the insertion tasks such as the geometric and spatial information of the cables and receptacles. To enhance generality, we use a foundation model, Segment Anything Model 2 (SAM2). To eleminate human intervention, we employ a Vision-Language Model (VLM) to automate the initial prompting of SAM2 to find segmentation masks. In the experiments, our method exhibits zero-shot capabilities, which enable direct deployments to real environments without fine-tuning.

\end{abstract}\vspace{-3pt}

\def\abstractname{Note to Practitioners}
\begin{abstract}
Automating the insertion of flexible flat cables (FFCs) is particularly challenging owing to their deformable structure and the submillimeter precision required in real-world assembly tasks. Traditional methods relying on manual teaching are time-consuming, inconsistent, and not scalable for high-throughput manufacturing. In this work, we introduce a reinforcement learning framework trained entirely in simulation, avoiding the risks and costs associated with real-world exploration. By leveraging foundation models in vision and language, our system automatically identifies task-relevant features and enables zero-shot deployment to real robots without any fine-tuning. This approach provides a practical, generalizable, and scalable solution for industrial insertion tasks—without manual labeling, task-specific engineering, or hardware-specific customization.

\end{abstract}


\section{Introduction}
\begin{figure}[t]
\vspace{-5pt}
\captionsetup{skip=0pt}
\centering
    \begin{subfigure}{0.34\textwidth}
   \captionsetup{skip=0pt}
	\includegraphics[width=\textwidth]{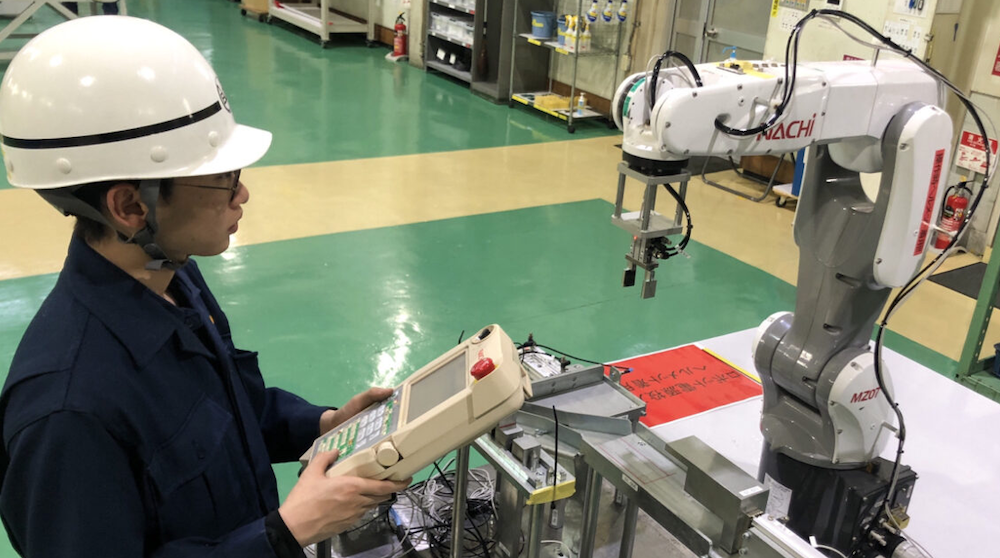}
	\caption{A human worker teaching a 6-DOF manipulator}
    \label{fig:ex_robot}
  \end{subfigure}
    \begin{subfigure}{0.34\textwidth}
    \captionsetup{skip=0pt}
        \includegraphics[width=\textwidth]{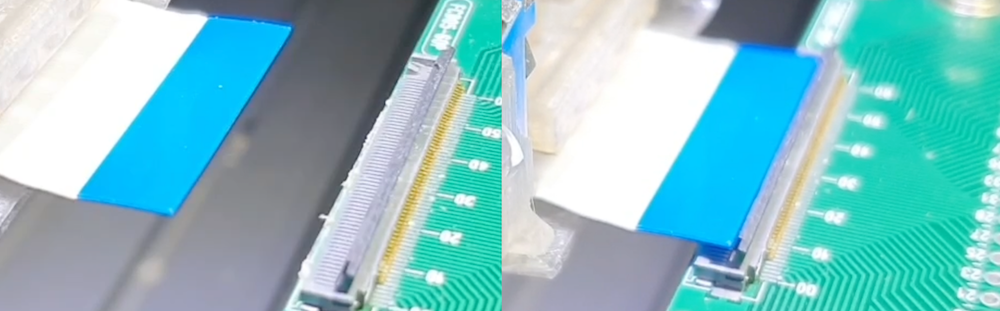}
        \caption{An FFC being inserted into a receptacle}
        \label{fig:ex_fpc}
    \end{subfigure}
    \begin{subfigure}{0.34\textwidth}
    \captionsetup{skip=0pt}
        \includegraphics[width=\textwidth]{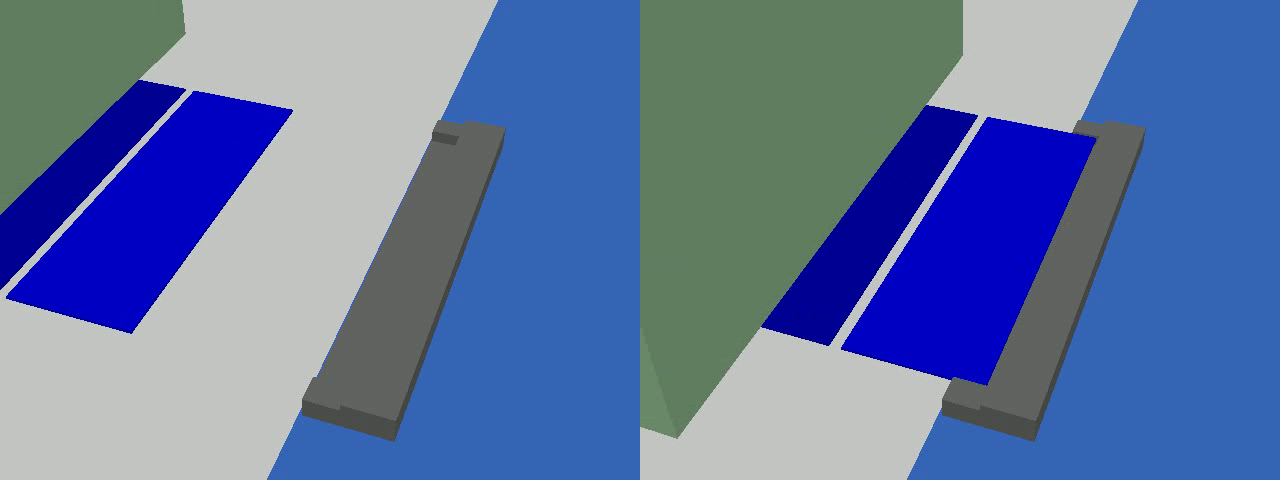}
        \caption{A visually approximated insertion task}
        \label{fig:ex_fpc_sim}
    \end{subfigure}
    \caption{An industrial cable insertion requiring a submillimeter precision}
        \label{fig:ex}
  \vspace{-15pt}
\end{figure}

\IEEEPARstart{T}{he} industrial insertion of flexible flat cables (FFCs) into receptacles is a critical yet challenging task in manufacturing which demands submillimeter precision while handling the deformable cables (Fig.~\ref{fig:ex}). This process is integral to numerous manufacturing operations, especially in the production of electronic devices with displays. The task is further complicated by the deformable nature of the cable, which can unpredictably bend upon contact with obstacles.
Traditionally, FFC insertion has relied heavily on human-guided trajectory generation for manipulators (Fig.~\ref{fig:ex_robot}), a labor-intensive approach that is both time-consuming and prone to error. Reinforcement Learning (RL) can offer the possibility to learn the optimal policy in an end-to-end manner without explicit modeling of the dynamics of the deformable soft body, which is often unavailable or prohibitively expensive. Nevertheless, the nondeterminism caused by the deformable FFCs requires significant training time. Moreover, training directly in a real environment is dangerous as industrial robots move fast and possess no safety measure so can damage the robots and nearby facilities.

In this paper, we propose an RL algorithm that enables robust FFC insertion by training an agent on a visually approximated task and deploying it to the real world using real-to-sim transfer. As shown in Figs.~\ref{fig:ex_fpc} and \ref{fig:ex_fpc_sim}, the visually approximated insertion task is constructed in a simulated environment by removing unnecessary visual information of objects such as colors, shades, and textures while preserving the geometric and spatial information relevant to the task such as shapes and their areas. Training is performed on this approximated task so that the RL agent can learn the insertion task without the complications caused by noisy and varying visual observations and safety concerns.
For the trained agent to be deployed to the real environment without fine-tuning, we develop a real-to-sim transfer approach, based on a foundation model Segment Anything Model 2 (SAM2)~\cite{sam2}. It  transforms the real-world insertion task into the visually approximated task. Thus, the RL agent performs the same task in the real world by using its learned knowledge in simulation. 

Even with the real-to-sim approach, generalization to various real environments remains limited since the task approximation cannot be done uniformly across every real-world instance. For example, the camera views and the sizes of FFCs and receptacles can vary in the real environments. To bridge this gap, we apply domain randomization in training, varying the camera extrinsic parameters (i.e., position, orientation, and field of view) and the dimensions of the FFC and receptacle.

The contributions of this work include:
\begin{itemize}
    \item \textbf{Visual approximation of task}: The use of the visually approximated task in simulation enables the RL agent to learn task-relevant features only, which are necessary for insertion. Our real-to-sim approach done by the visual approximation in the real world enables zero-shot transfer as the RL agent can perform the learned task even under the significant changes of the environment. In addition, training only in simulation removes the risk of physical damages of robots and surroundings caused by the initial random exploration.
    \item \textbf{Automated segmentation}: While foundation models such as SAM2 increase generality of our method by predicting segmentation masks of various cables and receptacles, they require human intervention to specify what to segment. To fully automate the entire process, we develop a technique generating prompts for SAM2 using  a VLM. 
    \item \textbf{Real-world deployment}: Our method successfully accomplishes the FFC insertion task in the real world using a 6-DOF industrial manipulator (Nachi MZ07L) equipped with two cameras. Our experimental result demonstrates that  difficult manufacturing processes which still rely heavily on human labor can be done autonomously while achieving faster completion time compared to the current technologies.

\end{itemize}

\section{Related Work}
\vspace{-3pt}
There has been much effort on increasing sample efficiency and generalization performance. Many RL algorithm using visual observations have proposed to augment data to enhance the sample efficiency and performance (e.g., \cite{curl,rad,drq,drqv2}). Some work~\cite{curl, sacae} use auxiliary losses like contrastive learning or reconstruction loss from an autoencoder to accelerate the training of encoder networks for visual observations. However, it is claimed that using such auxiliary losses rather harms the performance~\cite{drq,drqv2}, and using data augmentation only can outperform human instructed task execution and converge faster than methods using auxiliary losses. Recently, RL benefits from a pretrained vision foundation model (e.g., \cite{dinov2,clip}) which can provide good representations. They prove generality and strong performance across diverse visual tasks, motivating their adoption in RL with pixel observations.

Recently, RL benefits from pretrained vision foundation models (e.g., DINOv2~\cite{dinov2}, CLIP~\cite{clip}) which can provide good representations. Several works demonstrate that these foundation models achieve generality and strong performance across diverse visual tasks, motivating their adoption in RL with pixel observations. For instance, R3M~\cite{r3m} provides pretrained visual representations for robot manipulation using diverse human video datasets, while VIP~\cite{vip} employs time-contrastive learning to capture temporal dependencies in videos for goal-conditioned value functions, and MVP~\cite{mvp} proposes to learn visual pre-trained models for imitation learning using masked autoencoders.

RL has been widely applied to insertion tasks, particularly in rigid-body settings such as USB or HDMI connectors~\cite{visualinputnaturalreward, shield, edriad, pearl}. Among these, model-based RL learns a dynamics model of the environment and is sample-efficient, showing strong performance even in long-term credit assignment or sparse reward tasks. However, model errors can accumulate over time, and it remains challenging to predict the next state of deformable objects due to their complex dynamics.

To improve learning efficiency and safety, several studies incorporate prior knowledge. Works that utilize human demonstrations (e.g., \cite{ddpgfd,shield,visualinputnaturalreward,pbDbasedInsertion,rLfD}) accelerate policy learning and help avoid collisions. Residual RL~\cite{residual} combines classical control with RL by treating RL actions as corrections to planned trajectories, and some approaches~\cite{LfDRRRD} combine both demonstrations and residual RL for enhanced robustness. However, most of these focus on rigid-body insertion, where dynamics are known or simplified. Only a few studies explore soft-body insertion~\cite{ddpgfd,edriad}, and these often require additional modules such as vision or motion capture systems.

Reward design also plays a crucial role in insertion tasks. Some works construct dense reward functions using image or state information, or employ classifiers to determine goal achievement~\cite{raq,drem,industreal}. Nevertheless, these methods often rely on rigid-body assumptions or access to accurate state information.

In terms of sim-to-real transfer, \cite{residual} and \cite{pearl} require real-world fine-tuning after training in simulation, while \cite{nodynamicsrandomization} avoids dynamics randomization to enhance robustness against real-world perturbations, albeit using only state inputs without complex visual data. \cite{real2sim} leverages GAN-based unpaired image translation to bridge the visual gap between simulation and reality. \cite{learningfromSAM} uses a vision encoder with segmentation masks from the SAM for better generalization. More recently, \cite{nerf2real} employs NeRF to reconstruct 3D environments and address both view variance and zero-shot sim-to-real transfer, though it requires camera extrinsic and intrinsic parameters. However, most existing approaches either require extensive real-world fine-tuning, rely on high-fidelity simulation models, or are limited to rigid-body manipulation tasks.

Despite these advances, significant gaps remain in the literature for deformable object manipulation in industrial settings. Current methods for flexible cable insertion typically require either precise dynamics modeling or extensive real-world training data, both of which are challenging to obtain for industrial FFC insertion tasks. Furthermore, existing sim-to-real approaches often struggle with the visual complexity and precision requirements of manufacturing environments while maintaining safety during exploration. Our work addresses these limitations by proposing a visually approximated training approach combined with foundation model-based real-to-sim transfer, enabling zero-shot deployment for flexible cable insertion tasks without the risks associated with real-world training.

\section{Preliminaries}
\label{sec:prob}
\vspace{-3pt}

In this section, we present the background of the methods used in our work to better describe our proposed method.

\subsection{Reinforcement learning}
\vspace{-3pt}

A Markov Decision Process is defined as a tuple $(\mathcal{S}, \mathcal{A}, \mathcal{P}, \mathcal{}, \gamma)$ where  $\mathcal{S}$ and $\mathcal{A}$ denote finite sets of states and actions, respectively. Also, $\mathcal{R}: \mathcal{S} \times \mathcal{A} \times \mathcal{S} \rightarrow \mathbb{R}$ is a reward function, $\mathcal{P}: \mathcal{S} \times \mathcal{A} \times \mathcal{S} \rightarrow [0, 1]$ is a state-transition function, and $\gamma \in [0, 1]$ is the discount factor. The goal of  RL is to find the optimal policy that maximizes cumulative reward through iterative explorations of the state space.

\subsection{Semantic segmentation}
\vspace{-3pt}

Foundation models trained on large-scale datasets have demonstrated remarkable generalization across tasks. VLMs can perform a range of vision-based tasks without task-specific tuning. Among these, Segment Anything Model (SAM)~\cite{learningfromSAM} provides a promptable segmentation framework that outputs object masks based on points, boxes, or textual inputs. Its successor, SAM2, extends this functionality to video and real-time applications~\cite{sam2realtime}. SAM-based approaches offer rapid deployment and general-purpose flexibility, especially useful in deformable object manipulation where training data is scarce.

\subsection{Sim-to-real and real-to-sim techniques}
\vspace{-3pt}

Simulation environments enable safe and efficient training of RL, particularly for robotic tasks. However, transferring policies to the real world remains challenging owing to the reality gap. To address this, sim-to-real and real-to-sim strategies have been studied, typically classified into system specification, domain adaptation, and domain randomization.

System specification aims to replicate the real environment with photorealistic rendering and accurate physics models. While this can narrow the gap, it is computationally expensive and often infeasible, especially for complex or deformable dynamics. Domain adaptation transfers knowledge learned in simulation to real domains through approaches such as domain translation, policy distillation, or transfer learning. 
Domain randomization, in contrast, enhances robustness by exposing agents to wide variations in simulated textures, colors, lighting, and dynamics. This technique reduces reliance on accurate real-world modeling and is widely used due to its simplicity and effectiveness~\cite{nodynamicsrandomization,real2sim}.

\section{Problem Description}

We formulate this task as a Markov Decision Process (MDP) defined by the tuple $(\mathcal{S}, \mathcal{A}, \mathcal{P}, \mathcal{R}, \gamma)$. The state representation consists of visual observations and proprioceptive information. Visual observations are obtained from two cameras mounted on the end-effector of a 6-DOF manipulator, capturing complementary viewpoints that implicitly encode 3D geometric structure of the task space. Instead of using raw RGB images, we employ semantic segmentation masks that preserve only the geometric and spatial information of the FFC and receptacle while removing irrelevant visual information such as textures, colors, and background. The proprioceptive component is the orientation of the end-effector. Formally, the state at time $t$ is represented as:
\begin{equation}
s_t = \{M^{(1)}_t, M^{(2)}_t, \mathbf{q}_t\}
\end{equation}
where $M^{(1)}_t$ and $M^{(2)}_t$ are the segmentation masks from the two cameras, 
$\mathbf{q}_t \in SO(3)$ represents the end-effector orientation.
The action space consists of 6-dimensional continuous pose commands where each action $a_t \in \mathbb{R}^6$ is defined as:
\begin{equation}
a_t = [\Delta x, \Delta y, \Delta z, \Delta \phi, \Delta \theta, \Delta \psi]^T
\end{equation}
where $[\Delta x, \Delta y, \Delta z]$ represents translational displacements and $[\Delta \phi, \Delta \theta, \Delta \psi]$ represents rotational displacements about the end-effector coordinate axes. 

The objective is to learn an optimal policy $\pi^*$ that maximizes the expected cumulative reward over time, guiding the robot to complete the insertion task successfully. A discount factor $\gamma \in [0, 1]$ is used to balance immediate and future rewards.

\section{Methods}

\begin{figure*}[t]
\vspace{-5pt}
\captionsetup{skip=0pt}
    \centering
   \includegraphics[width=0.82\textwidth]{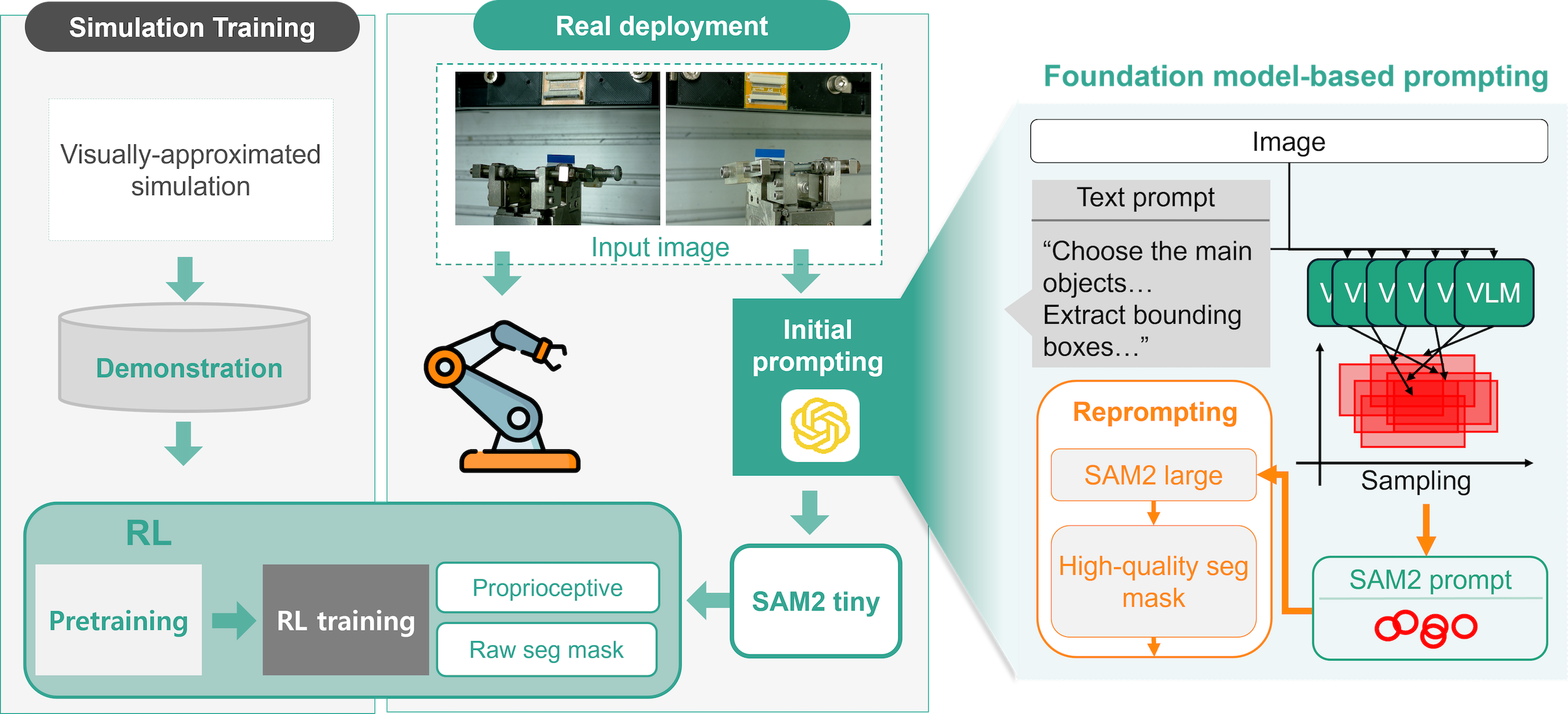}
    \caption{The architecture of the proposed framework, predefined prompt for VLM will generate prompt points for SAM2, leading to raw segmentation masks of the scene for observation of the RL agent}
    \label{fig:framework}\vspace{0pt}
\end{figure*}

Our approach trains an RL agent in simulation only but can achieve zero-shot transfer to real environments. The basic idea is to construct a visually approximated task for training which preserves the task-relevant information. When this agent is deployed, our real-to-sim approach transforms the real setting into the approximated one so that the agent can performs the same task that is learned during training. For a fully autonomous pipeline with general performance, we use foundation models that can generate the approximated task across different types and sizes of FFCs and receptacles. The entire framework of the proposed method is presented in Fig.~\ref{fig:framework}.

\subsection{Training on the visually approximated insertion task}
\label{sec:alg}

RL for physical robots faces inherent challenges due to the initial random exploration required, which can result in catastrophic failures, including physical damage to robots and their surroundings. Simulation has been widely adopted to mitigate such risks by visually and physically replicating task environments. However, creating high-fidelity simulations, especially for deformable objects with complex dynamics and unpredictable interactions, is computationally demanding. Another critical challenge arises when deploying a trained agent into the real world. Even a high-fidelity simulated environment rarely matches the exact distribution of the real world, necessitating sim-to-real transfer techniques. Fine-tuning in real-world settings is a common approach, yet it often demands extensive retraining, especially when significant discrepancies exist between simulation and reality.

A key lesson from our preliminary experiments is that detailed visual features from RGB images (e.g., textures, edges, and patterns) do not effectively support generalization of RL agents performing industrial insertion tasks. Since the visual details vary depending on FFC types, lighting conditions, robots, and camera configurations, it is impractical to run a comprehensive training across all possible variations.

To address these challenges, we develop a visually approximated version of the original FFC insertion task that includes only the essential visual information. 
Our method leverages semantic segmentation of RGB images to extract only geometric and spatial information (i.e., shapes and locations) of the cable and receptacle. To avoid environment-specific overfitting, robotic arms, grippers, and background features, which differ across manufacturing facilities, are simplified using segmentation masks.

\begin{figure}[t]
\vspace{-5pt}
\captionsetup{skip=0pt}
    \centering
  \begin{subfigure}{0.16\textwidth}
   \captionsetup{skip=0pt}
	\includegraphics[width=\textwidth]{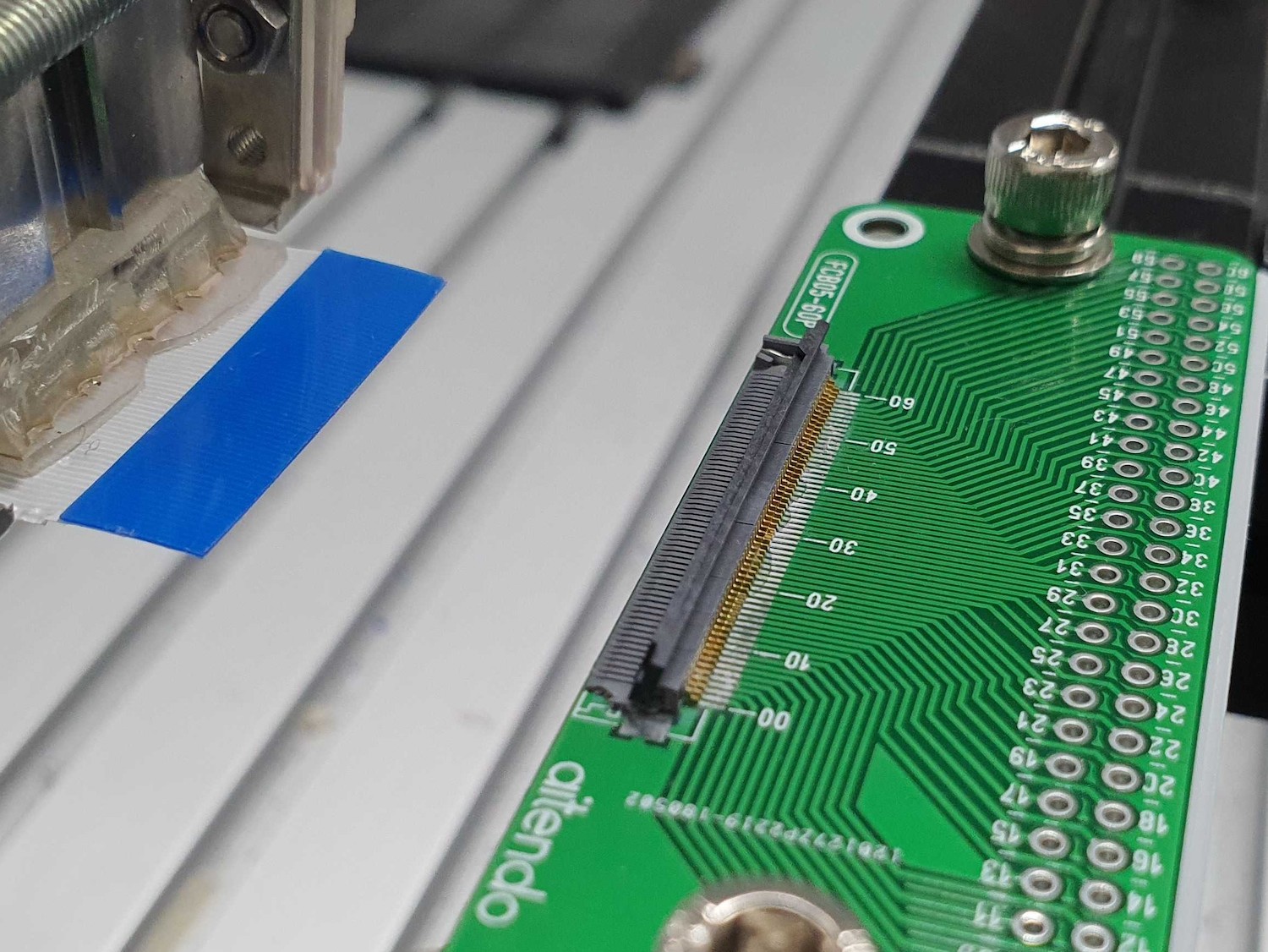}
	\caption{Original task}
    	\label{fig:vis_real}
  \end{subfigure}\,%
   \begin{subfigure}{0.16\textwidth}
   \captionsetup{skip=0pt}
	\includegraphics[width=\textwidth]{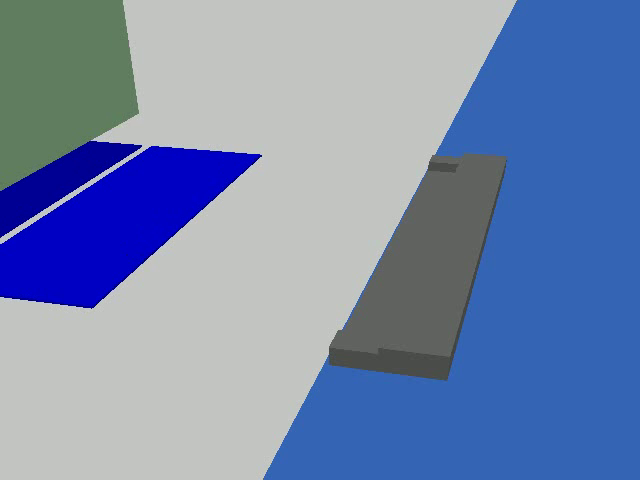}
	\caption{Simulation}
   	 \label{fig:vis_app}
  \end{subfigure}\,%
     \begin{subfigure}{0.16\textwidth}
   \captionsetup{skip=0pt}
	\includegraphics[width=\textwidth]{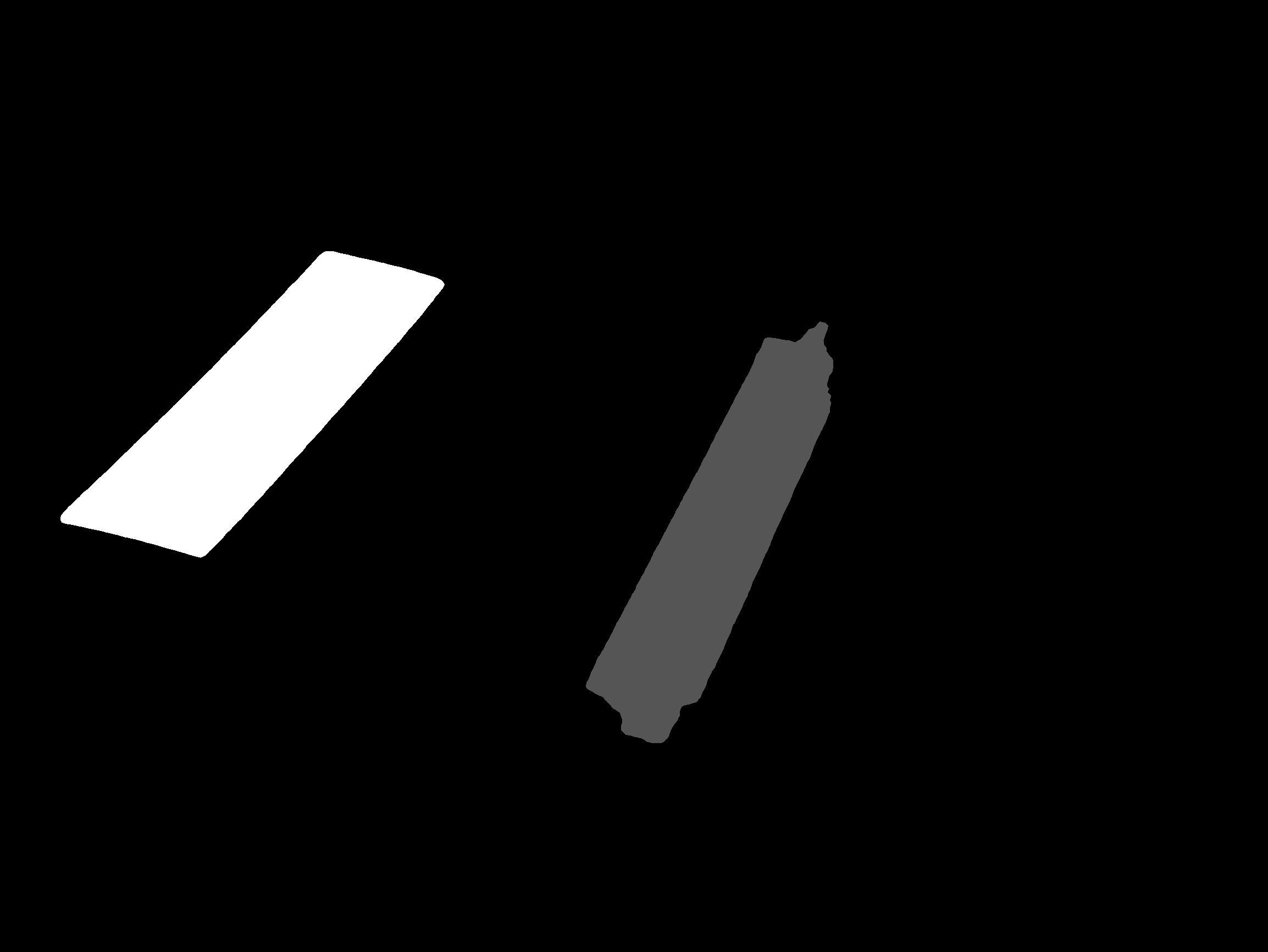}
	\caption{Real environment}
   	 \label{fig:vis_seg}
  \end{subfigure}
  \caption{An example of the visually approximated task. (a) shows the original image from the real environment. (b) shows the visually approximated task in simulation (PyBullet) which preserves task-relevant information only. (c) is the objects in the real environment represented by segmentation masks.} 
  \label{fig:vis}\vspace{-15pt}
\end{figure}

This visual approximation strategy is applied both in real-world and simulation environments as shown in Fig.~\ref{fig:vis}. In simulation (PyBullet is used), we model the FFC as a chain of thin hexahedrons, using joints between these segments to mimic cable elasticity (Fig.~\ref{fig:vis_app}). The FFC and receptacle are with solid colors resembling segmentation masks. In the real environment, segmentation masks for the FFC and receptacle are automatically generated using SAM2 as shown in Fig.~\ref{fig:vis_seg}. In this way, the visual gap between simulation and real-world environments is significantly reduced. Consequently, the RL agent learns policies focused solely on object-level interactions. Additionally, this approach is computationally efficient for training by eliminating the need for photorealistic rendering.

\subsection{The RL algorithm}
\label{sec:alg}

Off-policy RL is commonly used in robotics for its generalization capabilities and high sample efficiency from reusing experiences. Soft Actor-Critic (SAC)~\cite{sac} is one of the state-of-the-art off-policy algorithms with stable and fast convergence. SAC constructs objective functions based on an energy-based method to stabilize the training. It uses a policy network $\pi$ to produce $a_t \in \mathcal{A}$ in $s_t \in \mathcal{S}$ and has two $Q$ functions $Q_i$ with corresponding target networks $\hat{Q}_i$ for $i=1, 2$ to estimate the state-action value. The objective function to update the actor of SAC is:
\begin{equation}
\small{
\mathcal{J}(\pi)=\mathbb{E}_{s_t, a_t \sim \pi(s_t)}[\alpha\log(\pi(a_t|s_t))-\min_{i=1,2}{Q_i(s_t,a_t)}] \label{eqn:obj}
}
\end{equation}
The algorithm also updates the Q functions with:
\begin{equation}
\small{
\Hat{V}(s_{t+1})=\min_{i=1,2}{\Hat{Q_i}(s_{t+1},\pi(s_{t+1}))}-\alpha\log{\pi(a_{t+1}|s_{t+1})}
}
\end{equation}
\begin{equation}
\small{
\mathcal{J}(Q_i)=\frac{1}{2}\mathbb{E}_{s_t, a_t \sim \pi(s_t)}[(Q_i(s_t, a_t)-(r_t+\gamma\mathbb{E}_{s_{t+1}\sim \mathcal{P}}\Bar{V}(s_{t+1}))]
}
\end{equation}

Choosing the design of the reward function is a critical component of RL. A dense reward function provides frequent feedback, guiding the agent more easily toward the goal. However, it often requires domain knowledge and extensive tuning through trial and error. The performance is also sensitive to the choice and weighting of reward components. In contrast, a sparse reward function (or a natural reward) offers a simpler and more robust formulation, especially in real-world robotics where accurate state information (e.g., object poses) is difficult to obtain. As a result, sparse rewards are often preferred in practical robotic tasks~\cite{ddpgfd,edriad,shield,bcz}.

We employ a sparse reward structure to avoid the complexity of designing dense reward functions for deformable object manipulation. The reward function is defined as:
\begin{equation}
\small{
\mathcal{R}(s_t, a_t) = \begin{cases}
    10 & \text{if the FFC is successfully inserted} \\
    0  & \text{otherwise}
\end{cases}
}
\end{equation}

We construct a framework based on CoDER~\cite{coder}, an RL method designed for image-based manipulation tasks using sparse rewards. CoDER extends RAD~\cite{rad} by incorporating expert demonstrations and pretraining via a contrastive unsupervised loss (\ref{eqn:obj}). To improve training efficiency, we modify the original loss of CoDER by incorporating a regularized Q-function inspired by DrQ\cite{drq}. This modification introduces an additional gradient path from augmented observations alongside the standard Q-function gradient, significantly enhancing sample efficiency and accelerating convergence: 
\begin{equation}
\small{
y_t = r_t + \gamma \min_{j=1,2}\hat{Q}_j(s_{t+1}, a_{t+1}) - \alpha \log \pi(a_{t+1}|s_{t+1})
}
\label{eqn:target}
\end{equation}
where $a_{t+1} \sim \pi(\cdot|s_{t+1})$. The modified Q-function loss becomes:
\begin{equation}
\small{
\mathcal{J}(Q_i) = \mathbb{E}_{(s_t,a_t,r_t,s_{t+1}) \sim \mathcal{D}} \left[ (Q_i(s_t, a_t) - y_t)^2 + (Q_i(\tilde{s}_t, a_t) - y_t)^2 \right]
}
\label{eqn:obj_modified}
\end{equation}
where $\tilde{s}_t$ represents an augmented version of the state observation $s_t$. This regularization technique applies data augmentation to both the current state and the augmented state, effectively doubling the gradient signal and improving sample efficiency through implicit regularization of the Q-function.

\subsection{Dealing with view differences in training}
\label{sec:training}

While visual approximation through segmentation enables robust performance across variations in object size, color, and background, it remains sensitive to differences in camera viewpoints between training and deployment. Even minor changes in camera angle can significantly alter the perceived geometric and spatial relationships between the FFC and the receptacle. Since segmentation masks encode these spatial cues, any shift in perspective can result in inconsistent visual representations, widening the sim-to-real gap and potentially degrading policy performance.

To improve generalization across different hardware setups with varying camera configurations, we apply domain randomization during training. Specifically, we randomize the extrinsic parameters of the camera (i.e., position, orientation, and field of view) across training episodes. This exposes the policy to diverse viewpoints, encouraging it to learn viewpoint-invariant representations and enhancing robustness to camera variation at deployment time.

\subsection{Automated prompt generation for real-time segmentation}

\begin{figure*}
\vspace{-5pt}
\captionsetup{skip=0pt}
    \centering
   \includegraphics[width=0.95\textwidth]{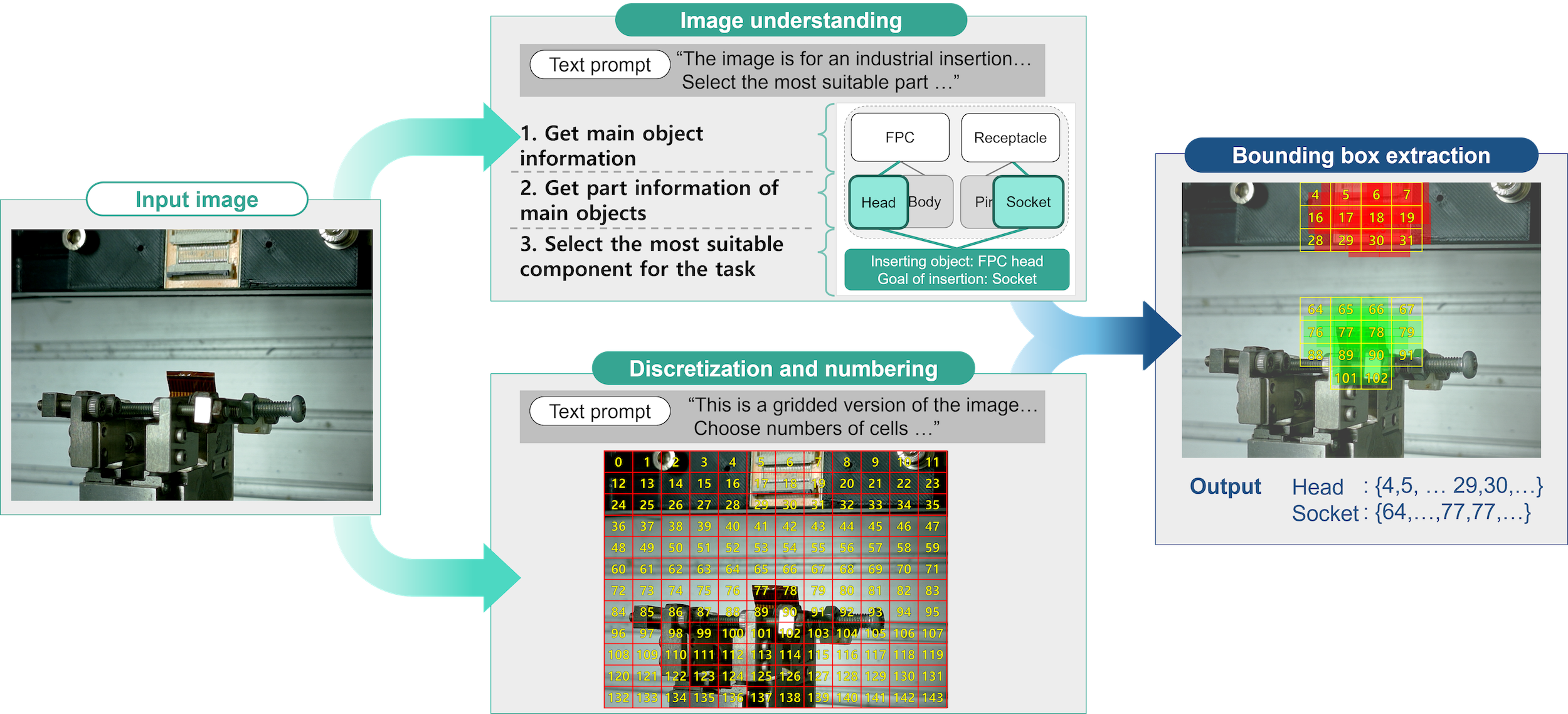}
    \caption{The automated pipeline for generating point prompts for SAM2. A VLM interprets a high-level task description, selects relevant regions from a grid discretizing the input image, and outputs cell indices. Overlapping red and green cells indicate high-confidence regions of the receptacle and FFC, respectively. An importance sampling is performed on the cells to obtain point prompts where the more overlaps incur higher probabilities of sampling.}
    \label{fig:flow_autoprompting}\vspace{0pt}
\end{figure*}

While our RL framework enables zero-shot insertion without fine-tuning, SAM2 still requires prompts to generate segmentation masks for the FFC and receptacle. These prompts are typically provided manually by clicking on target objects. To automate this process and enable real-time operation, we incorporate two key components: prompt automation using a VLM and prompt transfer for efficient inference. The overall automated pipeline is illustrated in Fig.~\ref{fig:flow_autoprompting}.

We use GPT-4o as the VLM, which receives a high-level task description that omits specific object attributes to preserve generality. While VLMs perform well in semantic reasoning, they lack pixel-level localization accuracy.\footnote{GPT-4o often failed to precisely identify bounding boxes for the FFC and receptacle.} To improve localization, we employ an in-context learning strategy combined with Monte Carlo sampling. The input image is discretized into a numbered spatial grid (bottom of Fig.~\ref{fig:flow_autoprompting}), framing the task as a multiple-choice question. We generate multiple outputs by varying the temperature between $0.0$ and $1.0$. As shown on the right side of Fig.~\ref{fig:flow_autoprompting}, grid cells with greater overlap across inference paths are more likely to contain objects and are thus sampled more frequently for prompt generation.

\begin{figure}[t]
\vspace{-5pt}
\captionsetup{skip=0pt}
    \captionsetup{skip=0pt}
    \centering
   \begin{subfigure}{0.22\textwidth}
   \captionsetup{skip=0pt}
	\includegraphics[width=\textwidth]{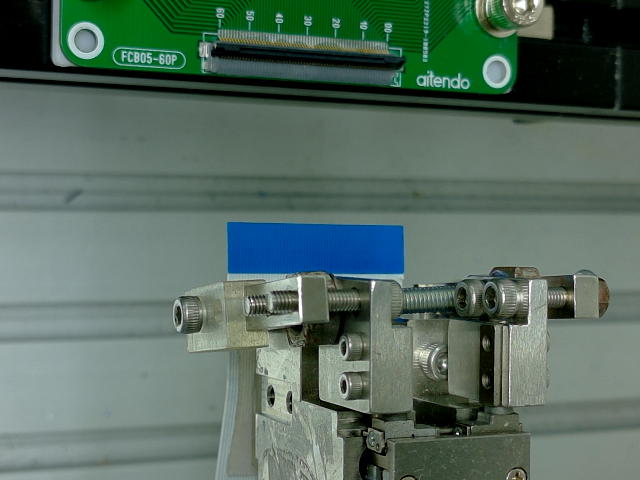}
	\caption{Raw image}
    \label{fig:raw}
  \end{subfigure}
  \begin{subfigure}{0.22\textwidth}
  \captionsetup{skip=0pt}
	\includegraphics[width=\textwidth]{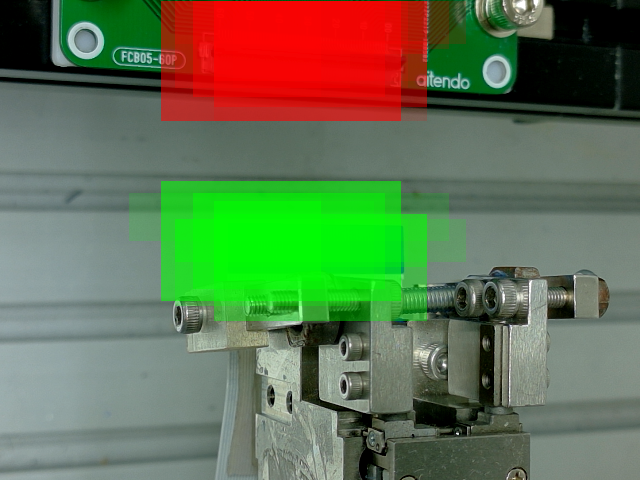}
	\caption{Bounding boxes}
    \label{fig:bb}
  \end{subfigure}
    \begin{subfigure}{0.22\textwidth}
  \captionsetup{skip=0pt}
	\includegraphics[width=\textwidth]{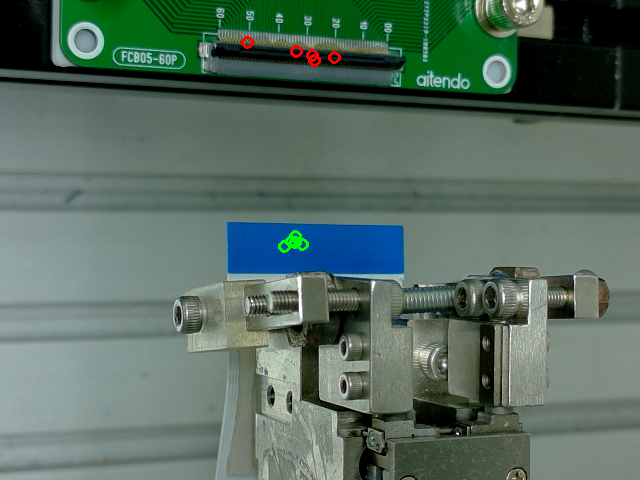}
	\caption{Sampled points}
    \label{fig:prompts}
  \end{subfigure}
    \begin{subfigure}{0.22\textwidth}
  \captionsetup{skip=0pt}
	\includegraphics[width=\textwidth]{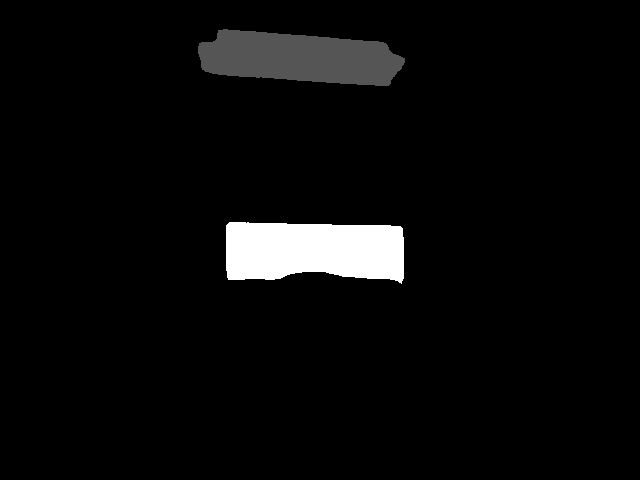}
	\caption{Resulting masks}
    \label{fig:masks}
  \end{subfigure}
  \caption{From the raw image in (a), the VLM performs multi-path Monte-Carlo inference to estimate bounding box distributions for the receptacle and FFC shown in (b). Prompt points in (c) are sampled via importance sampling  and used as inputs to SAM2, producing final segmentation masks in (d).}
  \label{fig:autoprompting_result}\vspace{-20pt}
\end{figure}

In the example shown in Fig.~\ref{fig:autoprompting_result}, we apply importance sampling to prioritize prompt points from high-overlap cells (Figs.~\ref{fig:bb} and \ref{fig:prompts}). These points are then used by SAM2 to generate segmentation masks for the target objects.

To further improve efficiency, we introduce a prompt transfer mechanism. We first run the largest version of SAM2 to obtain high-quality masks and then use those as prompts for a smaller, lightweight SAM2 model. This enables the smaller model to replicate the segmentation quality of the larger model at a significantly lower computational cost, supporting the low-latency demands of real-time robotic insertion.

\section{Experiments}
\label{sec:exp}
\vspace{-2pt}

In this section, we show the experimental results of the automated prompting method and the real-world deployment of the entire pipeline. All experiments are done in a system with AMD Ryzen 7 5800X CPU 3.80 GHz, NVIDIA GeForce RTX 3070ti GPU, and 32GB RAM. Training is done in the PyBullet simulation environment. As described in Sec.~\ref{sec:training}, domain randomization is applied to the camera extrinsic parameters to improve generalization. Specifically, the camera position is varied by $\pm 0.5$ cm in the $x$ and $y$ directions and $\pm 10$ cm in the $z$ direction. The camera yaw angle is randomized by $\pm 20^\circ$, and the field of view is sampled between $24^\circ$ and $50^\circ$. We also introduce random variations of $\pm 2$ cm in the $x$ and $y$ directions for the receptacle and FFC placement. The total number of training steps is $200,000$. Additional training hyperparameters are summarized in Table~\ref{tab:hyperparameter}.

\begin{table}[]
\vspace{15pt}
\centering
\caption{Hyperparameter setting for the training}
\scalebox{1}{%
\begin{tabular}{|c|c|}
\hline
Name & Value     \\ \hline
Optimizer & Adam  \\ \hline
Learning rate & 0.0001 for actor 0.001 for the others \\ \hline
Batch size & 128 \\ \hline
Buffer size & 100000  \\ \hline
Encoding dimension &  50  \\ \hline
Pretrain iteration & 1600  \\ \hline
Soft target update $\tau$ & 0.001  \\ \hline
Soft target update rate & 2  \\ \hline
\end{tabular}%
}
\label{tab:hyperparameter}
\vspace{-15pt}
\end{table}

\begin{figure}[t]
\vspace{-2pt}
\captionsetup{skip=0pt}
\centering
  \begin{subfigure}{0.39\textwidth}
        \captionsetup{skip=0pt}
	\includegraphics[width=\textwidth]{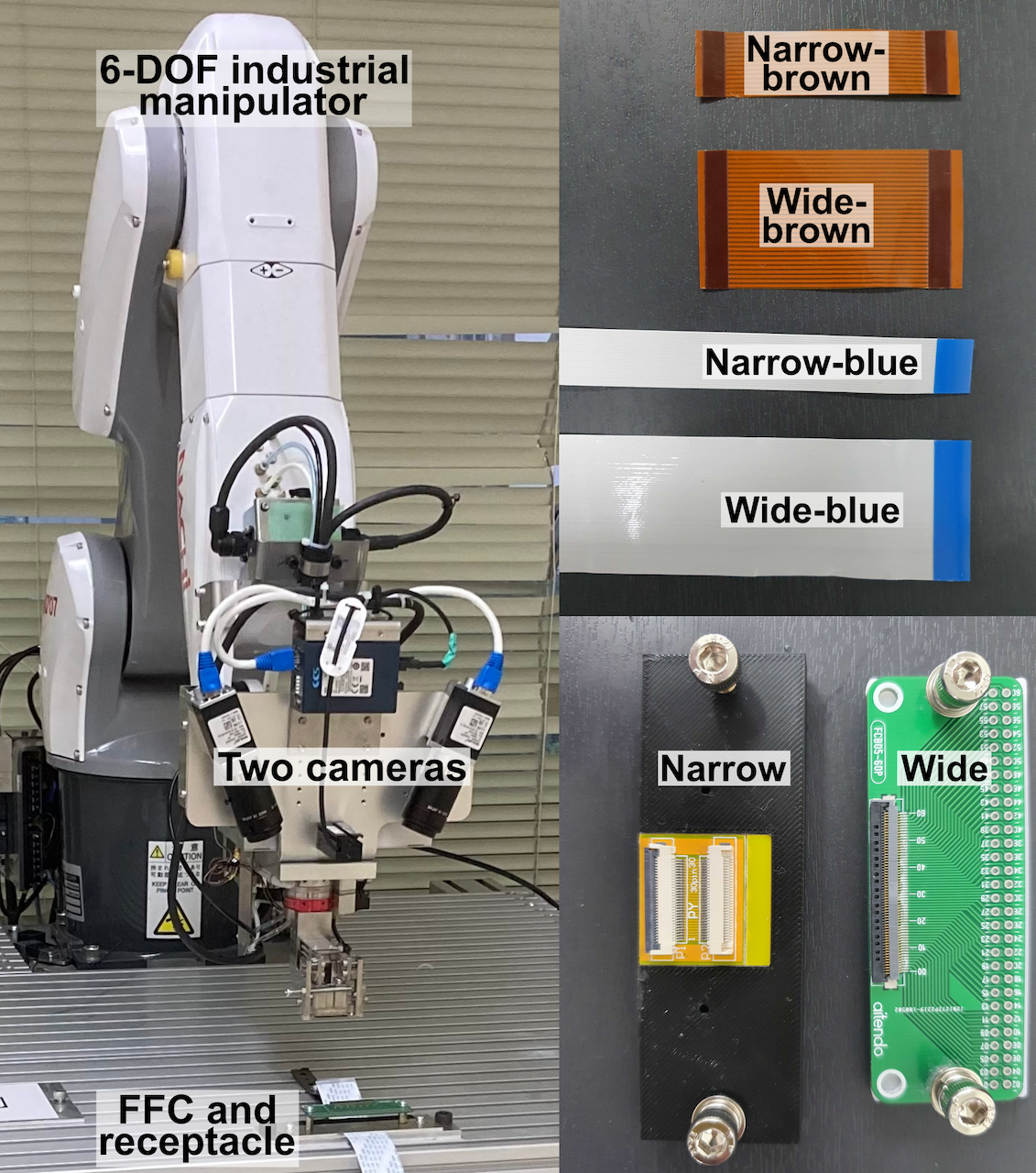}
        \caption{An industrial manipulator and FFCs/receptacles with varying colors and sizes}
        \label{fig:different types of objects}
  \end{subfigure}\\
  \begin{subfigure}{0.39\textwidth}
        \captionsetup{skip=0pt}
	\includegraphics[width=\textwidth]{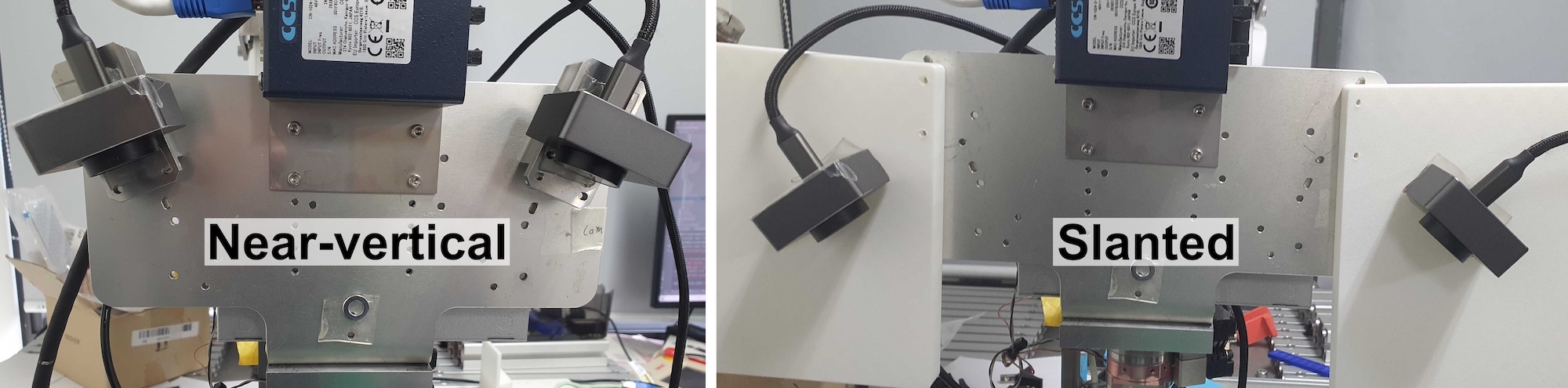}
	\caption{Two cameras capturing the FFC and receptacle downward}
    \label{fig:views}
  \end{subfigure}
  \caption{Real-world test environment. Our environment has  an industrial robot and different FFCs and receptacles shown in (a) where two cameras installed with different angles as described in (b).}
  \vspace{-10pt}
\end{figure}

We deploy our method in a real-world setup using a 6-DOF industrial manipulator (Nachi MZ-07), as shown on the left of Fig.~\ref{fig:different types of objects}. To evaluate generalization, we test the system with FFCs and receptacles of varying sizes (wide and narrow) and colors (blue and brown), as shown on the right of Fig.~\ref{fig:different types of objects}. The blue and brown FFCs differ in both color and surface pattern, while the receptacles vary substantially in shape, size, color, and texture. They also differ physically, with the brown ones being stiffer and the blue ones more flexible. For realism, we use commercially available receptacles without modification. These include protective covers that are not modeled in the simulation, introducing additional domain shifts. Such variations create realistic challenges in both perception and control, providing a strong benchmark for evaluating sim-to-real transfer and segmentation robustness. To assess performance under different viewpoints, we install 4K cameras (OBSBOT Meet 2) at two angles: near-vertical and slanted, as illustrated in Fig.~\ref{fig:views}.

\subsection{Automated prompt generation}
\label{sec:exp_prompt}

\begin{figure}[t]
\vspace{-2pt}
\captionsetup{skip=0pt}
\centering
  \begin{subfigure}{0.2\textwidth}
        \centering
        \captionsetup{skip=0pt}
	    \includegraphics[width=\textwidth]{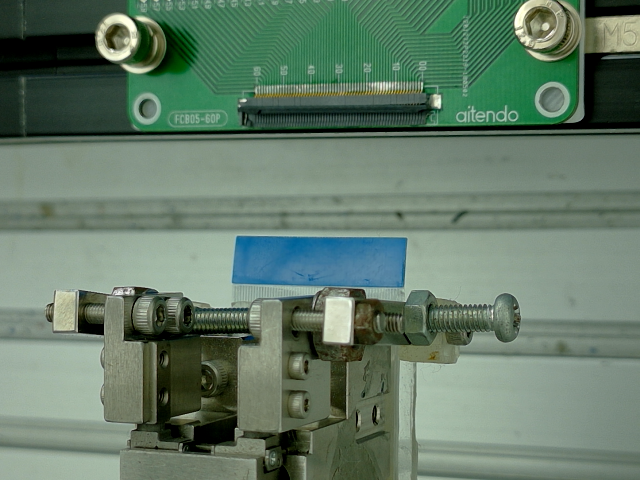}
	    \caption{Uniform lighting setting}
        \label{fig:light1}
  \end{subfigure}
  \begin{subfigure}{0.22\textwidth}
        \centering
        \captionsetup{skip=0pt}
	    \includegraphics[width=0.91\textwidth]{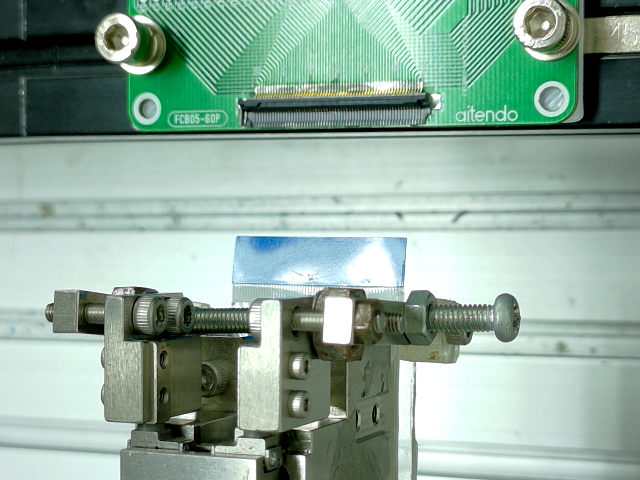}
        \caption{High-contrast lighting setting}
        \label{fig:light2}
  \end{subfigure}
  \caption{Example images under different lighting conditions. (a) Uniform lighting, which provides balanced illumination and is favorable for segmentation. (b) High-contrast lighting, which introduces strong shadows and reflections, making accurate segmentation more challenging.}
  \vspace{-5pt}
\end{figure}

We evaluate the robustness of VLM-based autoprompting by measuring the mean Intersection over Union (mIoU) between segmentation masks generated by our method and those obtained from manually prompted VLM outputs (used as the ground truth). Both our method and human provide five prompt points per image. Average mIoU scores are computed over 10 repetitions for each combination of FFC, receptacle, and lighting condition. To assess robustness under different illumination, we compare a uniform lighting setup (Fig.~\ref{fig:light1})---more favorable for segmentation---with a high-contrast setting (Fig.~\ref{fig:light2}), which introduces visual challenges. Each test image contains both an FFC and a receptacle, with wide FFCs always paired with wide receptacles to reflect realistic physical constraints. Table~\ref{tab:autopropmpting_performance} summarizes the results.

For the wide FFC and receptacle combinations, FFCs achieve high segmentation performance, with mIoU scores ranging from $97.02$\% to $99.93$\% across both lighting conditions. The brown FFC yields slightly lower scores than the blue FFC due to color ambiguity between brown and dark brown regions. In contrast, segmentation performance on the wide receptacles is generally lower---particularly under high-contrast lighting---with mIoUs ranging from $75.38$\% to $98.78$\%, primarily due to complex visual features such as embedded circuits and pin arrays. Overall, segmentation masks obtained under uniform illumination are consistently more accurate than those under high-contrast lighting.

For the narrow FFC and receptacle combinations, the trend remains consistent with the wide setting: receptacle segmentation generally yields lower mIoU scores than cable segmentation. However, unlike the wide setting, the mean mIoU for receptacles is notably lower. This gap arises from several factors. First, the smaller size of the receptacle makes it more difficult for the VLM to accurately select grid cells during autoprompting. Second, visual ambiguity increases when the receptacle’s cover is open, exposing an array of thin pins that complicate segmentation. In contrast, the narrow cables present fewer visual ambiguities, achieving up to $97.14$\% mIoU with small standard deviation.

\begin{table}[]
\centering
\caption{Mean mIoU scores and standard deviation for segmentation masks obtained from the autoprompting}
\scalebox{0.95}{%
\begin{tabular}{|c|c|c|c|}
\hline
Setting & Object    &  mIoU (\%)      & SD     \\ \hline
Wide-blue & FFC & 99.93 & 0.07 \\ \cline{2-4}
Uniform & Receptacle & 98.78 & 00.53 \\ \hline
Wide-blue & FFC & 99.21 & 0.30 \\ \cline{2-4}
High-contrast & Receptacle & 75.38 & 8.48 \\ \hline
Wide-brown & FFC & 98.23 & 3.48 \\ \cline{2-4}
Uniform & Receptacle & 90.97 & 24.46 \\ \hline
Wide-brown & FFC  & 97.02 & 7.12 \\ \cline{2-4}
High-contrast & Receptacle & 90.52 & 11.07 \\ \hline
Narrow-blue & FFC & 97.14 & 0.18 \\ \cline{2-4}
Uniform & Receptacle & 46.28 & 0.33 \\ \hline
Narrow-blue & FFC & 96.68 & 0.04 \\ \cline{2-4}
High-contrast & Receptacle & 51.78 & 0.39 \\ \hline
Narrow-brown & FFC & 95.38 & 0.00 \\ \cline{2-4}
Uniform & Receptacle & 21.11 & 0.00 \\ \hline
Narrow-brown & FFC  & 96.86 & 0.05 \\ \cline{2-4}
High-contrast & Receptacle & 62.72 & 0.35 \\ \hline
\end{tabular}%
}
\label{tab:autopropmpting_performance}
\end{table}

\subsection{Zero-shot real-world deployment}

\begin{figure}[t]
\vspace{-5pt}
\captionsetup{skip=0pt}
\centering
  \begin{subfigure}{0.241\textwidth}
  \captionsetup{skip=0pt}
	\includegraphics[width=\textwidth]{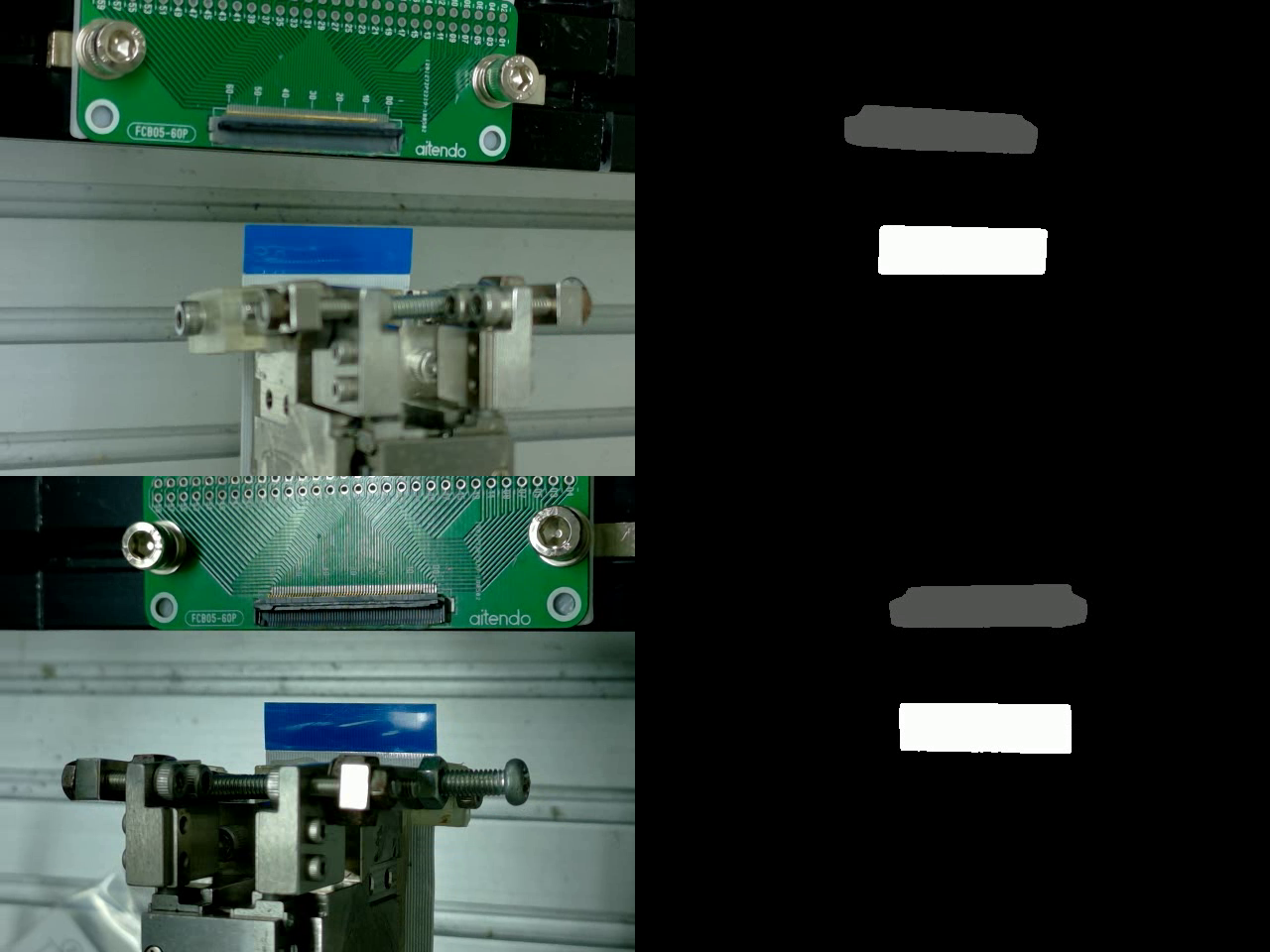}
	\caption{Near-vertical view}
    \label{fig:im1}
  \end{subfigure}
    \begin{subfigure}{0.241\textwidth}
  \captionsetup{skip=0pt}
	\includegraphics[width=\textwidth]{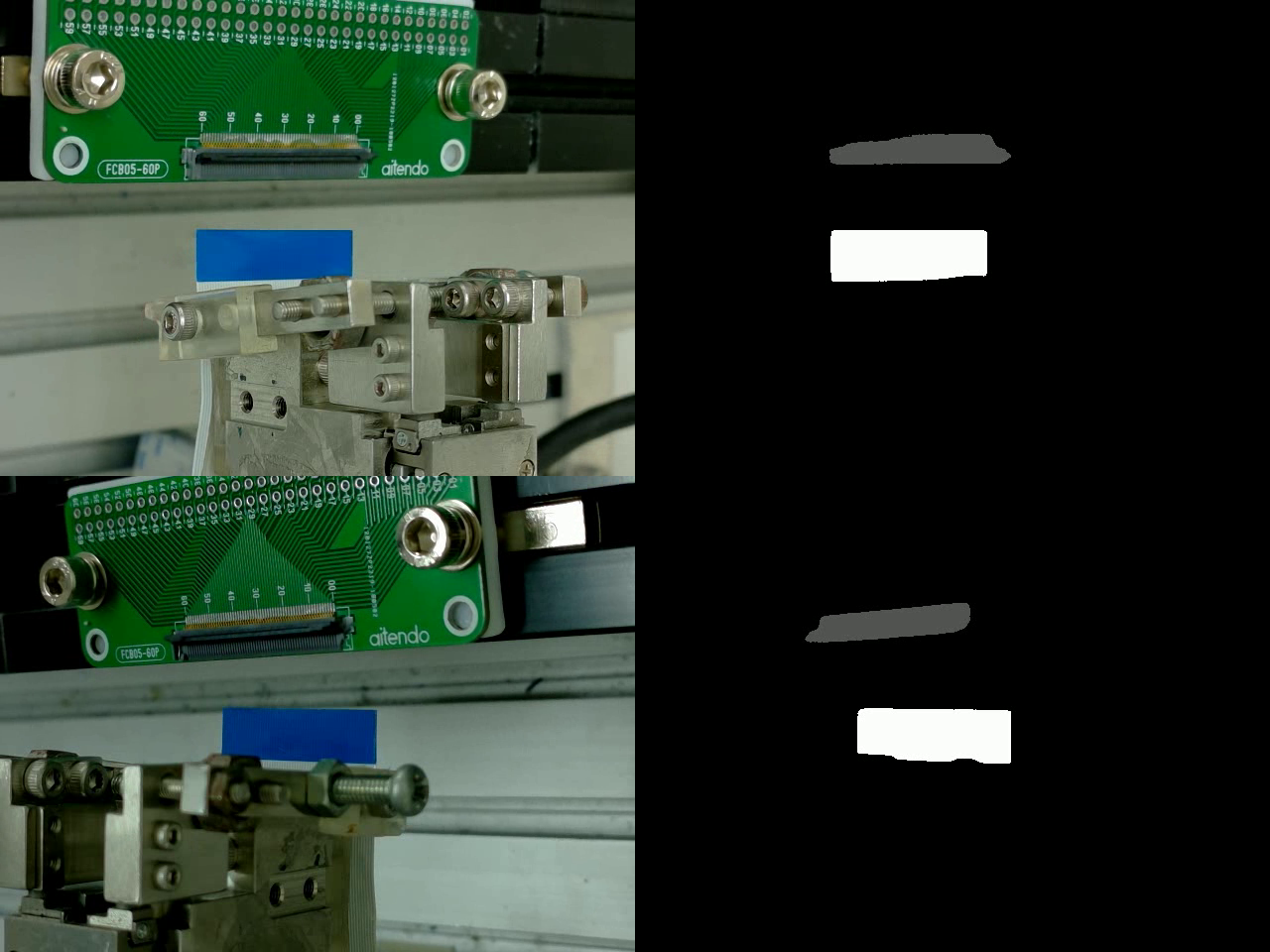}
	\caption{Slanted view}
    \label{fig:im2}
  \end{subfigure}
  \caption{Different camera views: (a) Near-vertical and (b) slanted perspectives, shown with corresponding RGB images and segmentation masks.}
  \label{fig:view variance}\vspace{-5pt}
\end{figure}

We evaluate zero-shot deployment over $10$ repetitions per setting under uniform lighting, which reflects typical manufacturing conditions. To assess the robustness of the proposed method under varying viewpoints, we compare two camera setups: a near-vertical view and a more steeply tilted slanted view as shown in Fig.~\ref{fig:view variance}. In every trial, the initial position of the robot end-effector is randomized within the camera frame. The base pose of the FFC tip is set $1.5$ cm away from the receptacle along the $x$-axis, aligned in the $y$- and $z$-axes, with an orientation of $(0^\circ, 0^\circ, 0^\circ)$. Around this base pose, the center of the FFC tip is randomized within a cuboid of $1$ cm ($x$), $0.5$ cm ($y$), and $1.5$ cm ($z$), and the orientation is perturbed within $\pm 7^\circ$ for roll, pitch, and yaw.

Table~\ref{tab:deployment} reports the success rates. An insertion is considered successful if the tip of the FFC is correctly placed onto the receptacle where both corners of the FFC remain within $1$ mm of the receptacle’s connection area, as illustrated in Fig.~\ref{fig:ex_fpc}. We also measure the average takt time, defined as the duration from the initial pose to successful insertion.

\begin{figure}[t]
\vspace{-5pt}
\captionsetup{skip=0pt}
\centering
  \begin{subfigure}{0.241\textwidth}
  \captionsetup{skip=0pt}
	\includegraphics[width=\textwidth]{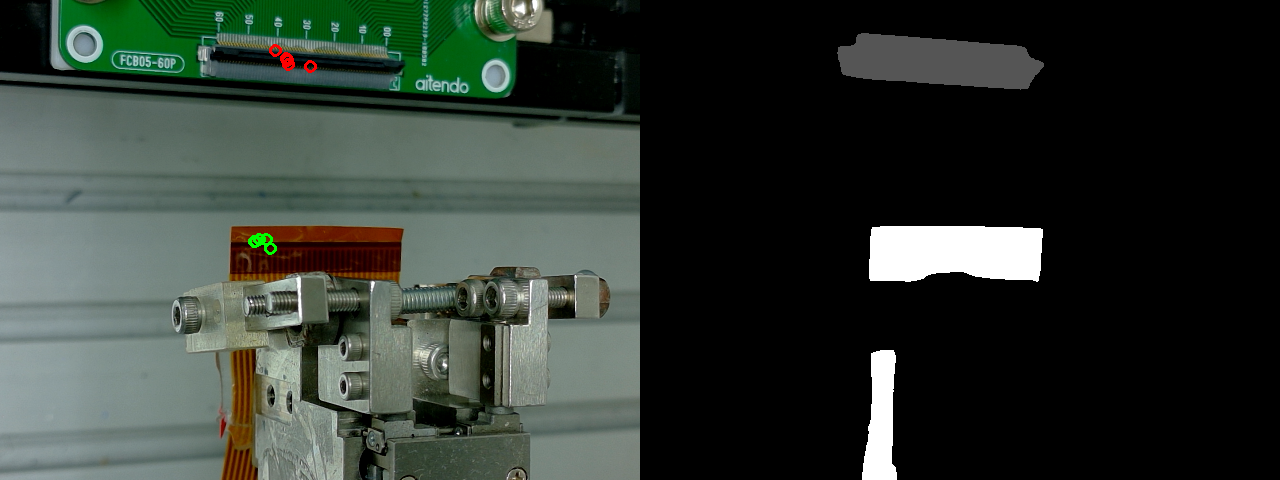}
	\caption{Wide-brown}
    \label{fig:exp_brown_wide}
  \end{subfigure}
    \begin{subfigure}{0.241\textwidth}
  \captionsetup{skip=0pt}
	\includegraphics[width=\textwidth]{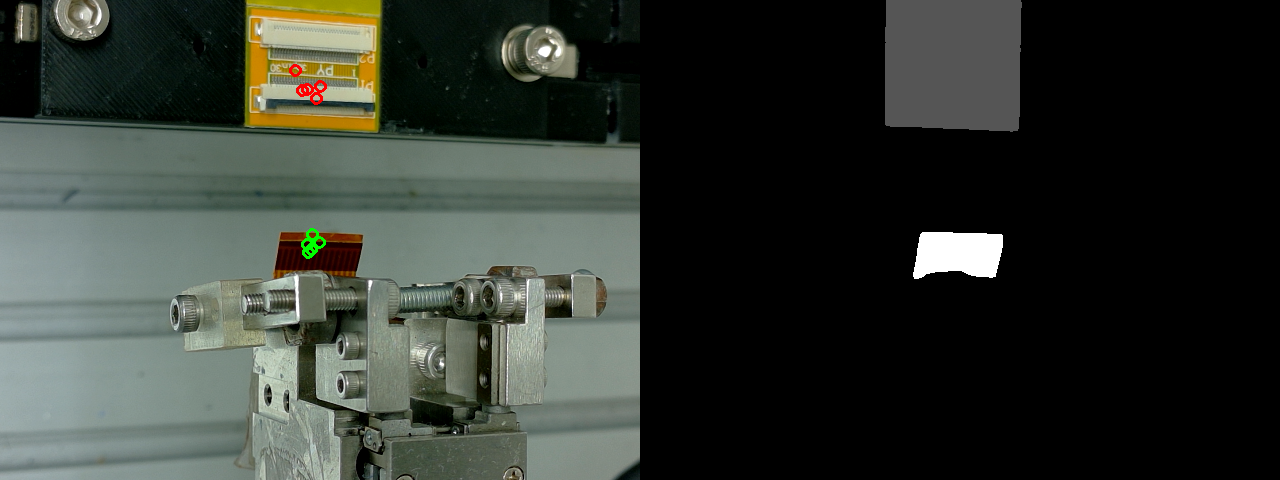}
	\caption{Narrow-brown}
    \label{fig:exp_brown_narrow}
  \end{subfigure}
  \caption{Segmentation masks of the brown FFCs. (a) The FFC body is recognized as part of the FFC head causing low success rates. (b) In the case of the narrow FFC, the incorrect segmentation does not occur as the body is fully occluded by the robot end-effector.}
  \label{fig:exp_brown}\vspace{-15pt}
\end{figure}

In the near-vertical camera view, the wide-blue FFC achieves a $90$\% success rate, while the wide-brown FFC reaches only $60$\% due to lower visual contrast and segmentation ambiguity. As shown in Fig.~\ref{fig:exp_brown_wide}, both the FFC head and part of body are included in the segmentation mask, leading to confusion. In contrast, the narrow-brown FFC achieves a higher success rate ($90$\%) because only the FFC head is visible (Fig.~\ref{fig:exp_brown_narrow}), preventing the body from being mistakenly segmented as part of the head. On successful trials, brown FFCs complete insertion faster on average ($22.17$ sec) than blue ones ($37.06$ sec). This difference is primarily due to their physical properties: the brown FFCs are stiffer, allowing for more stable insertions with fewer retries. In contrast, the more flexible blue FFCs often deform upon contact with the receptacle, requiring additional attempts to achieve successful alignment. In terms of the view, we find that the slanted view affects the performance negatively in general. In the segmentation mask of the receptacle in the slanted view, the opened cover of the wide receptacle makes a bump so changes the contour of the segmentation mask from the training environment.

In summary, our method achieves success rates of up to $90$\% without any fine-tuning, demonstrating strong zero-shot performance in industrial-grade FFC insertion tasks that demand submillimeter precision.

\begin{table}[]
\centering
\caption{Success rates and takt times for real-world deployment and view variance settings}
\scalebox{0.95}{%
\begin{tabular}{|c|c|c|c|c|c|}
\hline
View type & Size & Color & \begin{tabular}[c]{@{}c@{}}Success\\ rate (\%)\end{tabular} & \begin{tabular}[c]{@{}c@{}}Mean\\ time (sec) \end{tabular}& SD  \\ \hline
\multirow{4}{*}{Near-vertical} & \multirow{2}{*}{Wide}   & Blue & 90 & 38.40 & 13.74 \\ \cline{3-6}
                       &                        & Brown & 60 & 22.37 & 3.28  \\ \cline{2-6}
                       & \multirow{2}{*}{Narrow} & Blue & 70 & 35.71 & 17.72        \\ \cline{3-6}
                       &                        & Brown & 90 & 21.97 & 3.32  \\ \hline
\multirow{4}{*}{Slanted} & \multirow{2}{*}{Wide}   & Blue & 60 & 17.75 & 3.93  \\ \cline{3-6}
                       &                        & Brown & 60 & 33.25 & 15.79   \\ \cline{2-6}
                       & \multirow{2}{*}{Narrow} & Blue & 90 & 24.9 & 7.58   \\ \cline{3-6}
                       &                        & Brown & 90 & 19.25 & 3.26  \\ \hline
\end{tabular}%
}
\label{tab:deployment}

\end{table}

\section{Conclusion}
\vspace{-2pt}

In this work, we propose a reinforcement learning framework that achieves robust sim-to-real transfer for flexible flat cable (FFC) insertion by leveraging semantic segmentation. The approach isolates task-relevant geometric features while filtering out visual noise, enabling generalization across various real-world conditions without the need for fine-tuning. By incorporating segmentation foundation models, the method remains resilient to variations in camera viewpoints, object appearances, and background clutter. Future work will explore hybrid strategies that integrate complementary visual cues to extend applicability while maintaining the simplicity and generalization benefits of the proposed approach.

\bibliographystyle{IEEEtran}
\bibliography{references}

\end{document}